\pdfoutput=1

\documentclass[11pt]{article}

\usepackage[preprint]{acl}

\usepackage{times}
\usepackage{latexsym}

\usepackage[T1]{fontenc}

\usepackage[utf8]{inputenc}

\usepackage{microtype}

\usepackage{inconsolata}

\usepackage{graphicx}

\usepackage{amsmath,amssymb,amsfonts}
\usepackage{subcaption}
\usepackage{multirow}
\usepackage{graphicx}
\usepackage{pifont}
\usepackage{listings} 
\usepackage{xcolor}
\usepackage{newtxtext,newtxmath}
\usepackage{xcolor}
\usepackage{mdframed}
\usepackage{tabularx} 
\mdfsetup{%
linecolor=white,
backgroundcolor=gray!20,
}

\usepackage{authblk} 
\usepackage{hyperref} 

\title{Detection of LLM-Paraphrased Code and Identification of the Responsible LLM Using Coding Style Features}

\author{
\quad Shinwoo Park$^{1}$ 
\quad Hyundong Jin$^{1}$
\quad Jeong-won Cha$^{2}$ 
\quad Yo-Sub Han$^{\dagger, 1}$ \\
$^1$Yonsei University, Seoul, Republic of Korea \\
$^2$Changwon National University, Changwon-si, Gyeongsangnam-do, Republic of Korea \\
\texttt{\small pshkhh@yonsei.ac.kr, tuzi04@yonsei.ac.kr, jcha@gs.cwnu.ac.kr, emmous@yonsei.ac.kr} 
}

\newcommand{\correspondingfootnote}{
    \let\oldthefootnote=\thefootnote
    \renewcommand{\thefootnote}{}
    \footnotetext{$\dagger$ Corresponding author.}
    \let\thefootnote=\oldthefootnote
}

\begin{document}
\maketitle

\correspondingfootnote 

\newcommand{\eg}{{\it e.g.}}%
\newcommand{\ie}{{\it i.e.}}%

\begin{abstract}

Recent progress in large language models~(LLMs) for code generation has raised 
serious concerns about intellectual property protection. 
Malicious users can exploit LLMs to produce paraphrased versions of proprietary 
code that closely resemble the original. 
While the potential for LLM-assisted code paraphrasing continues to grow, 
research on detecting it remains limited, 
underscoring an urgent need for detection system.
We respond to this need by proposing two tasks. 
The first task is to detect whether code generated by an LLM 
is a paraphrased version of original human-written code.
The second task is to identify which LLM is used to paraphrase the original code. 
For these tasks, 
we construct a dataset \texttt{LPcode} consisting of 
pairs of human-written code and LLM-paraphrased code using various LLMs. 

We statistically confirm significant differences in the coding styles of 
human-written and LLM-paraphrased code, 
particularly in terms of naming consistency, code structure, and readability.
Based on these findings, 
we develop \texttt{LPcodedec}, a detection method that 
identifies paraphrase relationships between human-written and 
LLM-generated code, and discover which LLM is used for the paraphrasing.
\texttt{LPcodedec} outperforms the best baselines in two tasks, 
improving F1 scores by 2.64\% and 15.17\% while achieving speedups of 
1,343x and 213x, respectively.
Our code and data are available at \url{https://github.com/Shinwoo-Park/detecting_llm_paraphrased_code_via_coding_style_features}.

\end{abstract}

\section{Introduction}
\label{sec:introduction}

\begin{figure}[t!]
    \centering
    \begin{tabular}{c} 
       \begin{subfigure}[t]{\linewidth}
        \hspace{-5mm}\includegraphics[width=\columnwidth]{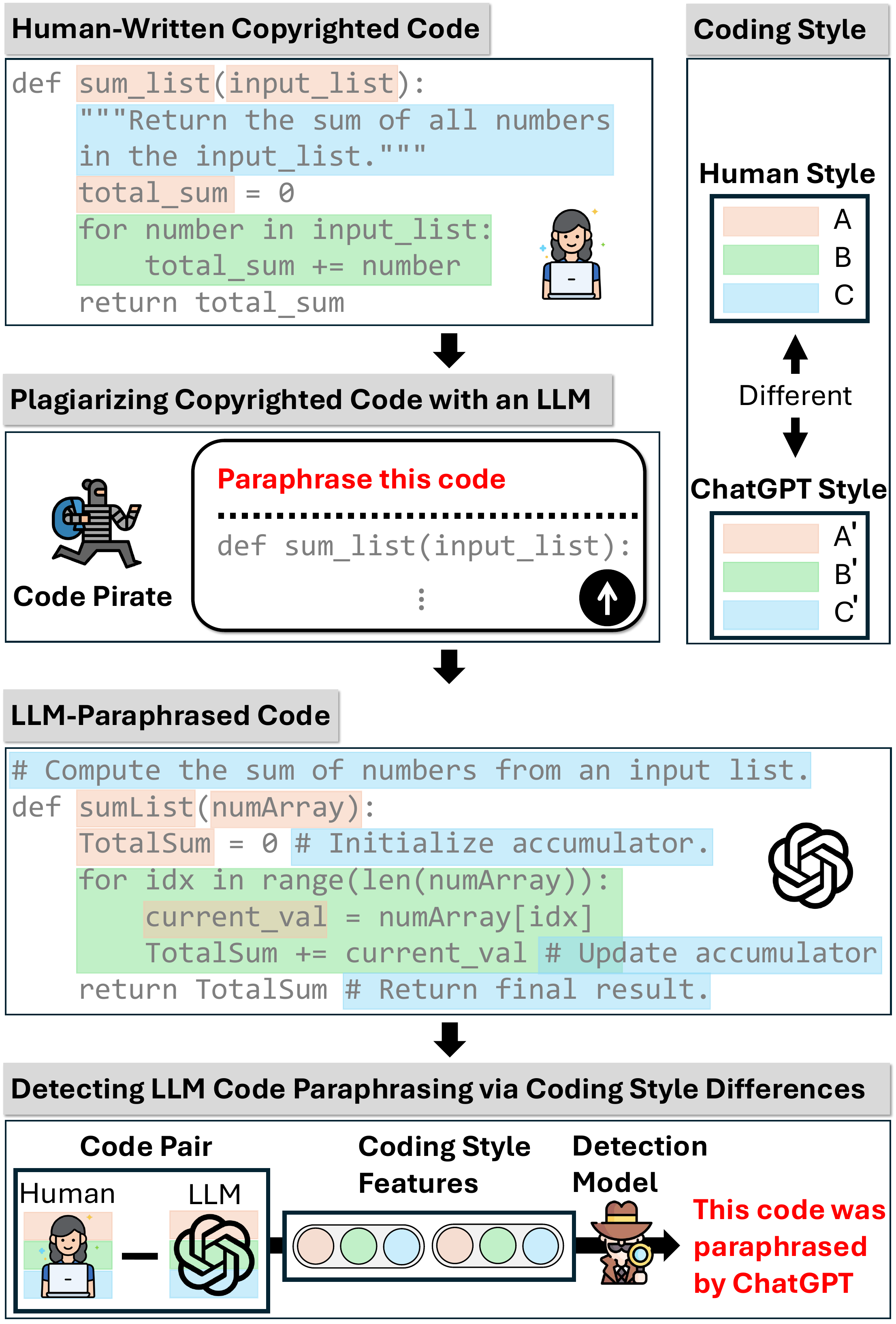}\hfill
            \centering
        \end{subfigure}\qquad
    \end{tabular}
    \caption{Illustration of LLM code paraphrasing detection using coding style.
    Humans and LLMs exhibit distinct patterns in naming, 
    structure, and comment usage when writing code.}\label{fig:motivation}
\end{figure}
Large language models have rapidly become a fundamental tool for developers, students, 
and researchers seeking automated solutions for 
code completion~\citep{izadi2024language,liu2024graphcoder,zhu2024exploring,cheng-etal-2024-dataflow}, 
generation~\citep{austin2021program,zheng2023codegeex,tong-zhang-2024-codejudge}, 
and translation~\citep{yuan2024transagent,he2025execoder,macedo2024intertrans}. 
Their increasing integration into software development pipelines
and educational environments has triggered serious concerns regarding 
unauthorized code reuse and potential plagiarism of copyrighted code. 
These concerns highlight the need not only to determine 
if a given code is generated by an LLM~\citep{nguyen2023snippet,yang2023zero,lee2024wrote,wang2023evaluating,shi2024detectcodegpt}, 
but also to verify whether it is a paraphrased version of an 
existing source~(for code plagiarism detection) and 
to track which LLM was used 
for the paraphrasing~(to enhance transparency in AI-assisted coding).

While paraphrase detection has been extensively studied in the field of 
natural language processing~\citep{wahle-etal-2022-large,krishna2023paraphrasing,tripto-etal-2024-ship}, 
to the best of our knowledge, 
no research has yet explored whether code generated by an LLM is a 
paraphrase of existing code or identified which LLM performed the paraphrasing.
We introduce the \texttt{LPcode}~(\textbf{L}LM-\textbf{P}araphrased \textbf{Code}) 
dataset to meet this new challenge. 
We first gather source code from GitHub that predates the widespread availability of advanced LLMs,
ensuring that the material is purely human-generated. 
Then, we provide these human-written code samples
as input to four different LLMs, 
prompting them to paraphrase and generate new versions of the original code. 
This procedure yields a comprehensive collection of human and 
LLM-generated code pairs that 
reflect various paraphrasing styles. 

We propose a novel approach for detecting code paraphrasing performed by LLMs. 
This approach quantifies the coding styles of both human developers and LLMs 
through features designed across three key aspects: 
naming conventions, code structure, and readability. 
The feature set comprises 10 quantitative metrics, 
including naming consistency, indentation consistency, and comment ratio. 
These features capture stylistic changes introduced when an LLM paraphrases 
human-written code, enabling the detection of paraphrase relationships 
between code samples. 
Additionally, our method leverages the unique coding style fingerprints 
of different LLMs to classify which model performed the paraphrasing.
Figure~\ref{fig:motivation} illustrates our LLM code 
paraphrasing detection approach.

Using ANOVA statistical analysis, we confirm that the most significant 
difference between human-written code and LLM-paraphrased code lies 
in the comment ratio. 
Building on these findings, we introduce 
\texttt{LPcodedec}~(\textbf{L}LM-\textbf{P}araphrased \textbf{Code} \textbf{De}te\textbf{c}tion), 
a detection method that exploits coding style features. 
Experimental results show that \texttt{LPcodedec} outperforms 
the strongest baselines in both tasks, 
improving F1 scores by 2.64\% and 15.17\%, 
while also accelerating detection by 1,343× and 213×, respectively.

\section{Background and Problem Definition}
\label{sec:related_work}

We begin by introducing code clone detection, 
a technique for identifying functionally equivalent code. 
Then, we highlight the key differences between code clone detection 
and our research focus: detecting LLM-based code paraphrasing.

\subsection{Code Clone Detection}

Code clone detection~\citep{svajlenko2014towards,zhang2023challenging,alam2023gptclonebench,dou2024cc2vec,feng2024machine} 
has been an active area of research in software engineering, 
aiming to identify duplicated or highly similar code fragments in large codebases. 
Clones are typically classified into four types: 
Type-1~(exact duplicates), 
Type-2~(syntactically similar with renamed identifiers or minor modifications), 
Type-3~(syntactically modified with additions or deletions), and 
Type-4~(semantically similar but syntactically different). 
Code clone detection methods identify whether two pieces of code are clones 
by analyzing token similarity, structural similarity, and embedding similarity.

\subsection{Detecting LLM-based Code Paraphrasing}

Our research differs in purpose and approach from code clone detection. 
Code clone detection research focuses on identifying 
semantic equivalence in code by analyzing the structure of source code, 
syntax trees, or token similarities to determine code duplication. 
In contrast, our work concentrates on capturing subtle changes 
in coding style that occur when LLMs
paraphrase human code.
Unlike code clone detection methods that focus on 
functional and structural similarities, 
our approach prioritizes stylistic characteristics 
such as naming conventions and indentation patterns. 

\begin{table*}[hbt!]
\centering
\begin{tabular}{cccccc|c}

\hline
\noalign{\hrule height 0.8pt}
\texttt{LPcode} & Human & ChatGPT & Gemini-Pro & WizardCoder & DeepSeek-Coder & Total
\\
\hline
C & 457 & 457 & 457 & 457 & 457 & 2,285
\\
C++ & 385 & 385 & 385 & 385 & 385 & 1,925
\\ 
Java & 1,494 & 1,494 & 1,494 & 1,494 & 1,494 & 7,470
\\
Python & 1,935 & 1,935 & 1,935 & 1,935 & 1,935 & 9,675
\\
\hline 

Total & 4,271 & 4,271 & 4,271 & 4,271 & 4,271 & 21,355
\\
\hline
\noalign{\hrule height 0.8pt}
\end{tabular}
\caption{
\texttt{LPcode} consists of human-written code and their 
paraphrased versions generated by four LLMs. 
}
\label{tab:data_statistics}
\end{table*}

\section{Dataset Construction: \texttt{LPcode}}
\label{sec:dataset_construction}

This section is organized as follows: 
(1) Collecting human-written code; 
(2) Generating LLM-paraphrased code based on the collected data; 
(3) Filtering the dataset; and 
(4) Introducing two tasks based on the \texttt{LPcode}.
The figure detailing the \texttt{LPcode} dataset construction process, 
along with descriptions of each step and the corresponding changes in the number 
of code samples, is provided in Appendix~\ref{sec:appendix_data_construct}.

\subsection{Human-Written Code Collection}
\label{sec:human_code_collection}

Since the emergence of LLMs with advanced code understanding and 
generation capabilities, 
such as Copilot and ChatGPT, 
LLM-generated code may be partially mixed 
into human-written code~\citep{wang2023evaluating}. 
Therefore, we collect 
human-written code
by crawling code\footnote{We refer to the entire source code contained 
in a single source code file as code.}
from GitHub repositories created before the emergence of LLMs.
Specifically, we collect repositories created between 
January 1, 2019, at 00:00 and 
January 1, 2020, at 00:00\footnote{We consider GPT-3, released in June 2020, as the first LLM with code generation capabilities.}
, which contain code written 
in C, C++, Java, and Python.
Additionally, we remove code with licenses 
other than Apache, BSD, and MIT. 

\subsection{LLM-Paraphrased Code Construction}
\label{sec:LLM_code_construction}

Malicious users may exploit LLMs to paraphrase copyrighted code 
rather than copying it directly, 
allowing them to misuse it without authorization. 
Considering real-world scenarios where LLMs are used for code theft, 
we collect LLM-paraphrased versions of human-written code.
We construct LLM-paraphrased code by providing human-written code
and instructing the LLMs to paraphrase them.
We use the following LLMs: 
1) OpenAI GPT-3.5;
2) Google Gemini-Pro;
3) WizardCoder-33B~\citep{luo2023wizardcoder};
4) DeepSeek-Coder-33B~\citep{guo2024deepseek}. 
We provide each LLM with human-written code and instruct 
it to paraphrase the given code. 
The prompt we use can be found in 
Figure~\ref{fig:appendix_prompt}.

\subsection{Data Filtering and Cleaning} 
\label{sec:dataset_filtering_cleaning}

We compute the similarity between 
human-written code
and 
LLM-paraphrased code based on the 
longest common subsequence~(LCS) 
and then calculate percentiles of this similarity.
Among the code, 
those with similarity scores of the 75th percentile or higher 
are considered nearly identical and are removed.
After that, we keep only the code that can be parsed into an 
abstract syntax tree~(AST) to ensure dataset integrity.
For data anonymization, we remove email addresses, URLs, and phone numbers 
included in code using regular expression.
Table~\ref{tab:data_statistics} presents the data statistics of the 
collected human-written code and LLM-paraphrased code.

\begin{table*}[hbt!]
\centering
\begin{tabular}{l|p{12cm}}
\hline
Feature Group & Description \\ \hline
Naming Consistency 
    & Evaluates adherence to naming conventions in functions, variables, classes, 
    and constants. \\ \hline
Code Structure 
    & Analyzes how code is structured through indentation patterns, function length 
    and nesting depth. \\ \hline
Readability 
    & Assesses code readability based on comment ratio and average lengths of function and variable names. \\ \hline
\end{tabular}
\caption{Overview of the three coding style feature groups.
Each of the three coding style feature groups analyzes a different aspect of the code.
Descriptions of the features within each feature group 
are provided in Appendix~\ref{sec:appendix_ten_features}.
}
\label{tab:feature_groups}
\end{table*}

\subsection{Two Proposed Tasks}

\label{sec:task_definition}

\paragraph{Task~1: LLM Code Paraphrasing Detection}

Task~1 aims to detect whether the LLM-generated code 
is a paraphrased version of the human-written code when 
given a pair of human and LLM-generated code.
This task is a binary classification problem. 
Our dataset consists of positive pairs, where LLM-generated code 
is a paraphrased version of human-written code, and negative pairs, 
where LLM-generated code is not related to the human-written code.
We randomly select the same number of negative pairs as 
positive pairs from all possible negative pairs to maintain a 
1:1 ratio between positive and negative pairs.

\paragraph{Task~2: LLM Provenance Tracking}

Task~2 is a multi-class classification 
task---given a human-written code and its paraphrased version generated by an LLM, identify which 
LLM among ChatGPT, Gemini-Pro, WizardCoder, or DeepSeek-Coder performed 
the paraphrasing.
We report the dataset sizes for Task~1 and Task~2 in Appendix~\ref{sec:appendix_data_size}.

\section{Approach: \texttt{LPcodedec}}
\label{sec:feature_analysis}

This section is structured as follows: 
(1) Designing features to quantify the coding styles of 
humans and LLMs; 
(2) Verifying whether the designed features exhibit 
significant differences between different code generators 
using ANOVA analysis; and 
(3) Developing a detection method, \texttt{LPcodedec}, 
based on the designed features. 

\subsection{Coding Style Feature Design}

We design three main feature groups to analyze the coding styles 
exhibited by humans and four LLMs when writing code. 
Table~\ref{tab:feature_groups} presents an overview of the 
three coding style feature groups.
Then, we define 10 features that capture distinct aspects of coding style, 
categorized into three groups:

\paragraph{Naming Consistency}
The first group consists of four metrics that measure consistency in 
function, variable, class, and constant naming practices. 
We identify five major naming patterns: 
camelCase, snake\_case, PascalCase, UPPER\_SNAKE\_CASE, and 
Other~(any format not fitting the previous four). 
For each of the four entity types—functions, variables, classes, and constants—we 
compute the frequency of each naming pattern and then measure how often the most frequently 
used pattern appears. 

\paragraph{Code Structure}
The second group targets structural aspects of code, 
including indentation consistency, average function length, and average nesting depth. 
Indentation consistency is determined by identifying the most frequently used 
indentation size~(e.g, four spaces) and dividing the count of this most common pattern 
by the total number of indentation patterns observed. 
A higher ratio indicates more uniform indentation. 
Average function length is the mean number of lines per function, 
which serves as a straightforward measure of overall code chunk size. 
Average nesting depth gauges the complexity of 
control flow structures~(such as nested loops or conditional statements) 
by calculating the mean level of nested blocks. 

\paragraph{Readability}
The final group consists of three features related to 
descriptiveness: comment ratio, average function name length, 
and average variable name length.
Comment ratio is the proportion of comment lines to total lines of code, 
reflecting how often the developer~(or an LLM) 
provides descriptions of functionality. 
Average function name length and average variable name length both indicate the level 
of descriptive detail used in naming. 

The detailed calculation methods for the 10 coding style features are provided 
in Appendix~\ref{sec:appendix_ten_features}.
We focus on these 10 features since they 
represent diverse dimensions of coding style. 
Naming consistency reveal stylistic preferences and adherence to established patterns, 
code structure features shed light on function organization, 
and readability captures how code is documented.
Since LLMs modify these features when paraphrasing human-written code, 
we propose to detect LLM code paraphrasing by leveraging changes in these features.

\begin{table*}[hbt!]
\centering

\begin{tabular}{ll|cc} 

\hline
\noalign{\hrule height 0.8pt}
Programming Language & Coding Style Feature & F-statistic & P-value~($\alpha$ = 5e-02)
\\ 
\hline

\multirow{2}{*}{C} 
& Comment Ratio & 153.71 & 3e-34\textsuperscript{*}
\\ 
 & Function Length & 19.92 & 8e-06\textsuperscript{*}
\\

\hline 

\multirow{2}{*}{C++} 
& Comment Ratio & 166.02 & 2e-36\textsuperscript{*}
\\ 
 & Class Naming Consistency & 3.87 & 4e-02\textsuperscript{*}
\\

\hline 

\multirow{2}{*}{Java} 
& Comment Ratio & 689.31 & 2e-145\textsuperscript{*}
\\ 
 & Variable Naming Consistency & 21.12 & 4e-06\textsuperscript{*}
\\

\hline 

\multirow{2}{*}{Python} 
& Comment Ratio  & 574.48 & 2e-123\textsuperscript{*}
\\ 
 & Function Length  & 70.40 & 6e-17\textsuperscript{*}
\\

\hline
\noalign{\hrule height 0.8pt}

\end{tabular}
\caption{We perform an ANOVA analysis on coding style features 
to compare human-written code and LLM-paraphrased code groups. 
For each programming language, 
we identify the two most statistically significant 
distinguishing features. 
All features in the table are statistically significant 
at a 0.05 significance level.
}\label{tab:anova}
\end{table*}

\subsection{Validation of Feature Effectiveness}

We conduct an ANOVA~(analysis of variance) test to examine the 
statistical significance of the ten defined features in 
distinguishing between human-written code and its paraphrased versions generated by LLMs. 
In this analysis, code generated by four different LLMs is grouped into a single LLM category, 
resulting in a comparison between two groups: human-written code and LLM-generated code.
We treat the ten extracted feature values from the human-written code set and 
the LLM group~(comprising code from all four LLMs) as observations. 
The group factor consists of two levels: human and LLM. 
We apply one-way ANOVA to each feature to determine whether it 
exhibits a statistically significant difference in distribution between human-written code 
and LLM-generated code. 
From the ANOVA results, we observe both the F-statistic and p-value. 
If the p-value is below the significance threshold~($\alpha$ = 0.05), 
we reject the null hypothesis, 
concluding that the corresponding feature shows a statistically significant difference 
between human-written and LLM-generated code. 
Additionally, the magnitude of the F-statistic indicates the extent to which 
the between-group mean difference explains the overall variance in the sample, 
meaning that a higher F-statistic suggests a greater distinction between the two groups.

Table~\ref{tab:anova} presents the top two features 
with the highest F-statistic values for each programming language, 
along with their p-values. 
The ANOVA results show that comment ratio has the highest F-statistic across 
all four programming languages, 
while the feature with the second-highest F-statistic varies by language. 
The fact that comment ratio consistently exhibits the largest F-statistic indicates 
that LLMs tend to follow a structured and repetitive approach to 
commenting when modifying human-written code. 
In contrast, human-written code varies significantly in commenting styles 
depending on the developer. 
Some developers write minimal or no comments, while others provide detailed, 
line-by-line explanations.
At the language level, the key distinguishing feature 
differs: class naming consistency stand out in C++, 
which emphasizes object-oriented design; 
variable naming consistency in Java; and 
function length in both C, known for its procedural style, and 
Python, which favors concise and readable code.

\subsection{Detection Method}
\label{sec:method}

We propose \texttt{LPcodedec}, a machine learning-based detection method that 
leverages our designed coding style features. 
For both human-written and LLM-generated code, we extract 10 feature values 
and represent each as a 10-dimensional feature vector. 
Then, we concatenate these two vectors and obtain a 
single 20-dimensional feature vector.
Using this feature representation as input, 
we train a fully connected neural network on the following two tasks:
1) Task 1: Decide whether or not the LLM-generated code in a given 
(human-written code, LLM-generated code) pair 
is a paraphrase of the human-written code.
2) Task 2: Given a pair of human-written code and 
its paraphrased version generated by an LLM, 
identify which LLM model performed the paraphrasing among 4 LLMs.

\texttt{LPcodedec} 
has two key characteristics: 
1) Explainability: Features such as naming conventions, 
function length, and comment ratio are relevant elements 
that developers focus on when understanding code. 
Since these interpretable features are used as model inputs, 
the learning outcomes are easier to interpret and explain.
2) Efficiency: The model simply relies on ten numerical features 
instead of  
hundreds of thousands of tokens or 
large-scale embeddings. 
This makes the model highly efficient in terms of both training and 
inference speed.
Additionally, the feature extraction process is performed through 
simple static analysis~(e.g., AST parsing and string pattern analysis), 
eliminating the need for pre-training on large-scale data.

\section{Experimental Settings}
\label{sec:experimental_settings}

\subsection{Baselines}
\label{sec:experimental_settings_baselines}

\paragraph{Task~1}
We select six baseline methods to detect whether 
a given pair of human-written and LLM-generated code has a paraphrasing relationship.
These baselines measure code similarity from 
three different perspectives: 
token-based similarity, structural similarity, and embedding similarity. 
1) Levenshtein Edit Distance: Measures textual similarity by calculating the 
minimum number of insertions, deletions, and substitutions required to 
transform one code into another. 
2) Jaccard Similarity: Computes similarity based on the ratio of the 
intersection to the union of token sets extracted from the code. 
3) Tree Edit Distance: Parses the code into an AST and measures 
structural similarity by calculating the cost of node insertions, deletions, 
and substitutions between two ASTs. 
4) Code LLM Embedding Similarity: Uses an LLM to compute the cosine similarity 
between the embeddings of human-written and LLM-generated code. 
We employ Qwen2.5-Coder~\citep{hui2024qwen2} 32B. 
5) MOSS: A widely used code plagiarism detection system developed by 
Stanford, commonly applied in educational settings.
MOSS analyzes common substrings and token sequences between two pieces of 
code while ensuring robustness against modifications such as 
variable renaming, whitespace changes, and comment alterations.
6) TF-IDF: Utilizes TF-IDF vectors of code tokens to train a 
fully connected layer model. 
TF-IDF measures the importance of a token by considering its frequency 
within a specific code while also reflecting how common it is across all code.

\paragraph{Task~2}
We employ TF-IDF as the baseline for Task~2, which involves identifying 
the LLM that paraphrased a given human-written code. 
The reasons for choosing TF-IDF as the baseline is summarized in three points.
First, Task~2 presents a novel classification challenge that has not been 
explored in previous research. 
Since there is no established baseline model available, TF-IDF 
can serve as a practical starting point. 
Second, we introduce coding style features for this task. 
Comparing these features with TF-IDF, which relies on token 
frequency patterns, help assess the contribution of style-based 
characteristics in identifying the LLM responsible for paraphrasing. 
Third, LLMs often exhibit distinct token usage patterns depending on 
their training data and generation strategies.
TF-IDF captures these model-specific tendencies by measuring 
differences in token frequency distributions. 
Since the paraphrasing patterns of different LLMs can be reflected 
in the relative frequencies of tokens, 
TF-IDF serves as a reasonable and interpretable baseline for Task~2. 
By establishing how token frequency patterns vary across different LLMs, 
we can better understand whether these patterns alone provide 
sufficient signals for LLM identification.
In Appendix~\ref{sec:appendix_tfidf}, 
we analyze the differences in token usage patterns among LLMs.
We compute the TF-IDF vectors for 
code generated by different LLMs and compare 
the cosine similarity between these vectors to 
identify differences in token frequency patterns 
across models. 
The analysis reveals that ChatGPT exhibits distinct 
token frequency patterns compared to other LLMs.

\subsection{Evaluation Metrics}
\label{sec:experimental_settings_evaluation_metrics}
We use the execution time and the F1 score as our evaluation metrics. 
The execution time
measures the total processing time in seconds required to run each method 
from the data preprocessing to the model inference. 
The preprocessing step includes code tokenization~(for edit or Jaccard distance), 
AST parsing~(for tree edit distance), 
code embedding generation, 
TF-IDF computation, 
and extraction of 10 coding-style features. 
The training time for machine learning models is also recorded. 
Finally, the inference time for each code pair is summed into 
the overall execution cost. 
The F1 score is the harmonic mean of precision and recall. 

\subsection{Implementation Details}
\label{sec:appendix_implementation_details}

For experiments using MOSS, we use the submission script provided by Stanford.
We implement Jaccard Similarity and Levenshtein Edit Distance in Python, 
while Tree Edit Distance follows the method provided by \citet{song2024revisiting}.
We build the TF-IDF baseline and our proposed \texttt{LPcodedec} using scikit-learn, 
employing the MLPClassifier provided by scikit-learn.
We conduct our experiments on a server with an 
NVIDIA RTX A6000.

\begin{table*}[hbt!]
\centering\small
\begin{tabular}{l|cc|cc|cc|cc}
\hline
\noalign{\hrule height 0.8pt}

Task~1 & \multicolumn{2}{c|}{C} & \multicolumn{2}{c|}{C++}  & \multicolumn{2}{c|}{Java}  & \multicolumn{2}{c}{Python} 
\\
Detection Methods & Time & F1 & Time & F1 & Time & F1 & Time & F1
\\ 
\hline
LLM Embedding & 1210.76 & 63.00 & 983.26 & 66.36 & 3742.94 & 68.20 & 5328.64 & 50.78 
\\
TF-IDF & 148.49 & 20.75 & 95.10 & 14.98 & 1421.82 & 15.84 & 2001.44 & 35.77 
\\
Jaccard Similarity & \textbf{0.11} & 68.30 & \textbf{0.09} & 70.82 & \textbf{0.38} & 72.38 & \textbf{0.55} & 73.05 
\\
Levenshtein Edit Distance & 742.20 & 64.27 & 660.35 & 71.42 & 3942.18 & 71.42 & 5408.68 & 71.86 
\\
MOSS & 1261.32 & 66.81 & 1062.60 & 69.17 & 4123.44 & 72.37 & 5340.60 & 68.41  
\\
Tree Edit Distance & 1094.11 & \textbf{88.06} & 1641.17 & \underline{87.71} & 4892.87 & \underline{88.20} & 11523.28 & \underline{86.95}  
\\
\hline 
\texttt{LPcodedec} & \underline{1.59} & \underline{87.52} & \underline{1.31} & \textbf{88.39} & \underline{4.99} & \textbf{91.13} & \underline{6.37} & \textbf{93.16}

\\
\hline
\noalign{\hrule height 0.8pt}
\end{tabular}
\caption{Performance of Task~1.
We present the average time and F1-score across all folds of the 5-fold validation.
We highlight the best scores in bold and the second-best scores with an underline.
}
\label{tab:experimental_results_task1}
\end{table*}

\section{Results and Analysis}
\label{sec:experimental_results}

\subsection{LLM Code Paraphrasing Detection}
Table~\ref{tab:experimental_results_task1} presents the experimental results for Task~1.
We highlight the best performance in terms of execution time~(lower is better) and 
F1 score~(higher is better) in bold, while the second-best performance in underlined. 
On average across the four languages, 
\texttt{LPcodedec} achieves the highest F1 score, 
followed by Tree Edit Distance, 
while Jaccard Similarity is the fastest in execution time, 
with \texttt{LPcodedec} being the second fastest. 
Compared to Tree Edit Distance, \texttt{LPcodedec} achieves an average F1 score improvement of 2.64\% 
and is 1,343 times faster in execution. 
This demonstrates that our proposed method, which leverages coding style features, 
not only achieves superior performance but is also highly efficient.
The superior performance and fast execution time of \texttt{LPcodedec} demonstrate its ability 
to efficiently adapt to a wider range of programming languages and newly emerging LLMs with minimal cost.

\subsection{Comparison of the Three Feature Groups}

We investigate how each of three 
feature groups individually contributes to Task~1. 
We conduct a series of experiments which we train the same classification model 
using only one feature group at a time.
Table~\ref{tab:ablation_study} 
compares the performance of models trained for Task~1 
using only a single feature group~(naming consistency, code structure, or readability).
Our findings show that integrating three feature groups yields the highest overall 
F1 scores in all four
programming languages we examine. 
This result aligns with expectations, as each feature group captures a unique facet of coding style,
and merging them allows us to more comprehensively detect differences between human-written and LLM-generated code.

\begin{table}[hbt!]
\centering\small
\begin{tabular}{l|llll}
\hline
\noalign{\hrule height 0.8pt}

Feature Group & C & C++  & Java & Python
\\
\hline
All & \textbf{87.52} & \textbf{88.39} & \textbf{91.13} & \textbf{96.16 }
\\ 
\hline
Naming Consistency & 80.09 & 79.45 & 82.44 & 88.16 
\\ 
Code Structure & 80.57 & \underline{84.25} & \underline{85.47} & 87.17  
\\ 
Readability & \underline{83.55} & 83.05 & 85.17 & \underline{89.03} 

\\
\hline
\noalign{\hrule height 0.8pt}
\end{tabular}
\caption{Performance comparison of the three feature groups.
We present the average F1-score across all folds of the 5-fold validation.
We highlight the best scores in bold and the second-best scores with an underline.}
\label{tab:ablation_study}
\end{table}

When we focus on single feature groups, the best-performing group varies by language. 
For C and Python, the Readability group~(e.g., comment ratio, average function name length, average variable name length) 
produces the highest F1 scores among the three groups. 
For C, which frequently follows a procedural style and does not enforce formal structures like classes, 
developers often rely on personal conventions for naming and commenting. 
Some adopt very concise or even cryptic naming patterns and omit most comments, 
while others include detailed function-by-function annotations. 
By comparison, an LLM may introduce more systematic documentation or name variables and functions in a consistently 
recognizable pattern. 
As a result, features such as comment ratio and the 
lengths of function and variable names capture 
these stylistic gaps more effectively, 
making the Readability group particularly discriminative for C. 
Python, in particular, promotes simplicity and readability 
through official style guidelines like PEP 8. 
However, real-world developer practices often 
diverge significantly: some developers write extensive 
docstrings and comments, while others include minimal annotations. 
In contrast, many LLMs consistently 
generate structured docstrings and adopt standardized naming conventions, 
resulting in a pronounced gap 
in readability metrics between human-written and LLM-generated code.

Meanwhile, C++ and Java show their best single-group performance with the 
Code Structure features~(e.g., indentation consistency, average function length, average nesting depth). 
Both languages rely heavily on object-oriented principles, 
including classes and inheritance, encouraging complex but 
distinctive structural patterns. 
LLMs often generate code in a more standardized, textbook style, 
while human developers vary significantly in function segmentation, 
nesting style, and indentation usage. 
This leads to structural features serving as strong indicators of LLM-generated code in C++ and Java.

The different results for each individual feature group across languages offer practical insights into 
how LLM-generated code diverges from human code. 
In languages with substantial variability in readability practices~(like C and Python), 
metrics related to comments and naming length are particularly effective for detection. 
In contrast, in object-oriented languages such as C++ and Java, 
structural aspects become the primary differentiators. 
These observations highlight how varying dimensions of coding style contribute to detecting paraphrased code 
across programming languages.

\begin{table}[hbt!]
\centering\small
\begin{tabular}{l|c|c|c|c}
\hline
\noalign{\hrule height 0.8pt}

Task~2 & C & C++ & Java & Python 
\\
\hline
TF-IDF & 37.30 & 35.12 & 37.44 & 37.82 
\\
\texttt{LPcodedec} & \textbf{43.13} & \textbf{39.36} & \textbf{45.19} & \textbf{42.41}
\\
\hline
\noalign{\hrule height 0.8pt}
\end{tabular}
\caption{
Performance of Task~2.
We present the average F1-score across all folds of the 5-fold validation.
We highlight the best scores in bold.
}
\label{tab:experimental_results_task2}
\end{table}

\subsection{LLM Provenance Tracking}
Table~\ref{tab:experimental_results_task2} 
presents the experimental results for Task~2. 
Across the four languages, \texttt{LPcodedec} achieves an average F1 score that is 15.17\% higher 
than TF-IDF while also being 213 times faster in execution.
Despite training the same fully connected layers, the significant difference in execution speed is 
due to the TF-IDF feature vector being, on average, 2,660 times larger in dimension than the 
coding style feature vector used by \texttt{LPcodedec} across the four languages.
\texttt{LPcodedec} outperforms TF-IDF as it 
utilizes a wider range of coding style features, 
such as naming conventions, code structure, and 
readability patterns, 
instead of relying solely on token frequency patterns.
The experimental results demonstrate that considering 
various coding style features 
is effective for the task of identifying 
the LLM that paraphrased human-written code.

\subsection{Error Analysis}

\begin{figure}[hbt!]
    \centering
    \begin{tabular}{c} 
       \begin{subfigure}[t]{\linewidth}
         \hspace{-5mm}\includegraphics[width=6.5cm]{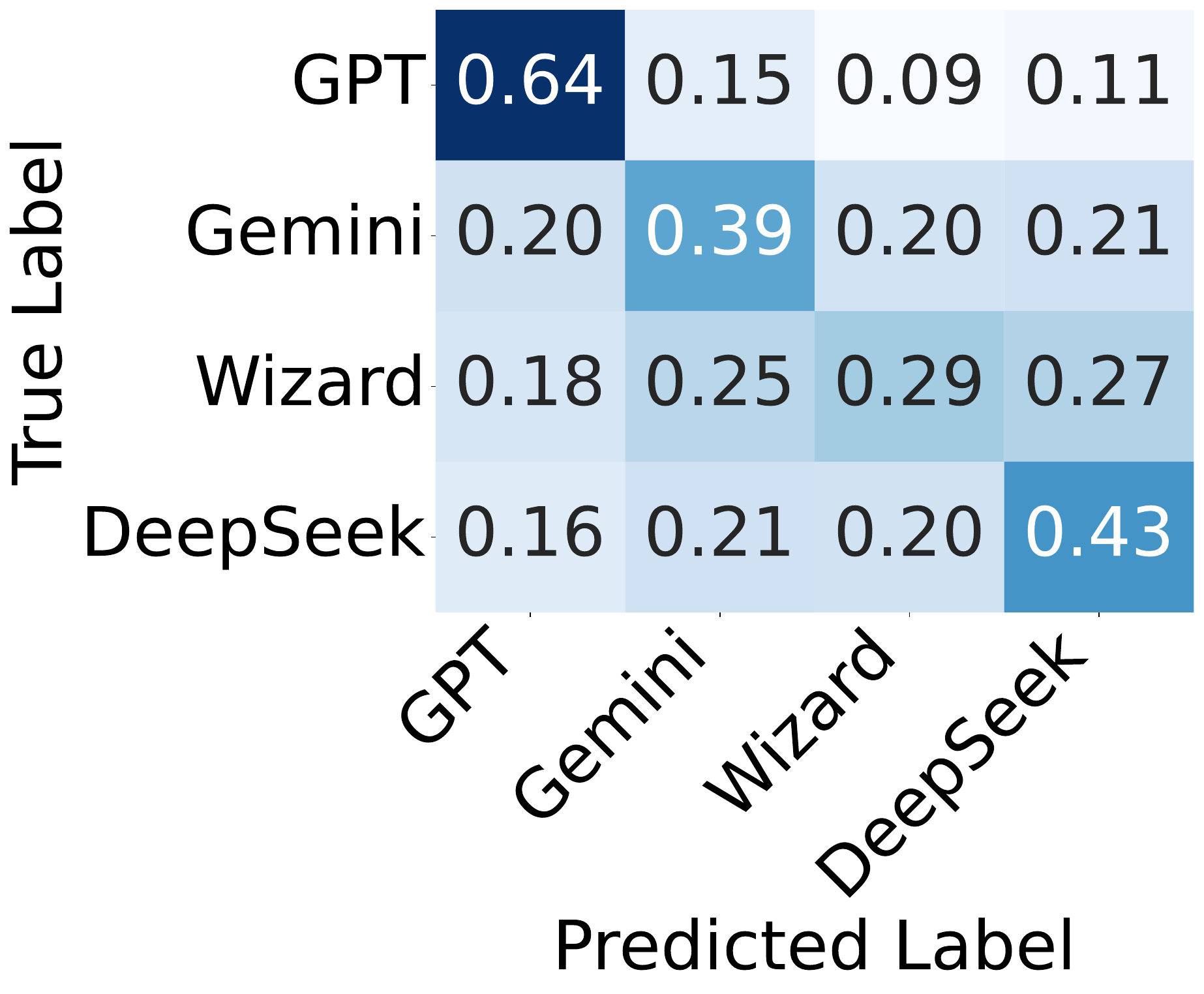}\hfill
            \centering
        \end{subfigure}\qquad
    \end{tabular}
    \caption{The confusion matrix showing the prediction results of \texttt{LPcodedec}.}\label{fig:error_analysis}
\end{figure}

We analyze the prediction failures of \texttt{LPcodedec} using a confusion matrix.
Figure~\ref{fig:error_analysis} shows the average \texttt{LPcodedec} predictions computed by first 
averaging across five folds for each of the four programming languages, 
and then averaging across the four languages.
We observe that \texttt{LPcodedec} performs best 
at identifying ChatGPT and struggles most with WizardCoder.
This indicates that ChatGPT has a more 
distinctive coding style and 
code paraphrasing approach compared to other LLMs, 
while WizardCoder may use a less clear or 
less consistent paraphrasing method. 

We find that \texttt{LPcodedec} most frequently misclassifies 
WizardCoder-generated code as DeepSeek-Coder-generated code, 
suggesting that the two LLMs may have similar coding styles.
In Appendix~\ref{sec:codebleu_ana}, 
we analyze the similarity between human-written code and code paraphrased by 
four LLMs using CodeBLEU~\citep{ren2020codebleu}. 
CodeBLEU is a metric that measures code similarity by comprehensively 
considering n-gram matches and AST similarity. 
Through this analysis, we find that the code generated 
by WizardCoder and DeepSeek-Coder exhibits the highest 
CodeBLEU similarity among the four LLMs. 
This high similarity indicates that these models produce 
code with closely aligned stylistic and structural patterns, 
which aligns with the observation that \texttt{LPcodedec} 
faces the greatest challenge in distinguishing between code 
from these two LLMs.

\section{Related Work}

As we propose the tasks of detecting and tracing LLM code paraphrasing, 
we have not found directly related works. 
Instead, we conducted a literature survey on detecting LLM-generated code, 
a binary classification task that determines whether a given code is human-written or 
LLM-generated.
\citet{lee2024wrote} divided the vocabulary in code generation through LLM 
into a Green/Red list and inserted detectable patterns in LLM-generated code by 
promoting the generation of tokens belonging to the Green list.
\citet{yang2023zero} proposed DetectGPT4Code, a modification of the existing 
machine-generated text detection method, 
DetectGPT~\citep{mitchell2023detectgpt}, using a code-specialized language model. 
\citet{wang2023evaluating} evaluated the effectiveness of existing 
machine-generated text detectors in detecting machine-generated code.
They reported that the existing detectors exhibit degraded performance 
in identifying machine-generated code in contrast to 
their performance in detecting machine-generated natural language text.
These studies focused on detecting LLM-generated code derived from 
natural language descriptions 
or function headers and did not address the detection of paraphrasing between 
human-written and LLM-generated code.
\citet{krishna2023paraphrasing} reported that applying paraphrasing to 
LLM-generated text can bypass existing detection methods. 
Therefore, studying the detection of LLM paraphrasing
is necessary to enhance the robustness of LLM-generated code detection methods.

\section{Conclusion}

The rapid advancement of LLMs in code understanding and 
generation raises an urgent need to protect copyrighted code 
and ensure transparent AI usage. 
In response to this challenge, we build \texttt{LPcode}, 
a dedicated resource to support research in this area. 
We introduce \texttt{LPcodedec}, 
a detection method that uses coding style features to 
identify whether a code sample has been paraphrased by an 
LLM and determine which LLM performed the paraphrasing. 
By analyzing the distinctive stylistic patterns in code, 
\texttt{LPcodedec} can be applied to tasks like 
academic plagiarism detection and open-source AI monitoring, 
helping prevent unauthorized code reuse and promote 
responsible AI practices.

\section*{Limitations}

While we introduce a novel 
dataset~(\texttt{LPcode}) and develop an
efficient detection method~(\texttt{LPcodedec}), 
several limitations highlight opportunities 
for future work.
First, our dataset and experiments focus on 
four programming languages and 
four specific LLMs. 
In reality, the range of programming languages and 
LLM variants is vast and continuously evolving. 
Extending \texttt{LPcode} to cover more languages 
and a broader set of LLMs would enhance the 
robustness and generalizability of our findings. 
Second, although our method demonstrates 
strong performance in 
Task~1~(determining whether LLM-generated code 
is a paraphrased version of human-written code), 
its performance is lower in Task~2, 
where the goal is to identify which specific LLM 
produced the paraphrased code.
Designing more refined features can enable 
the development of a model that captures 
subtle coding style differences among LLMs.
Finally, adversarial methods that randomize or 
obfuscate code style can potentially circumvent 
style-based detection, 
highlighting the need for ongoing research into 
complementary detection strategies.

\section*{Ethical Considerations}

We address pressing concerns related to code plagiarism, unauthorized usage, and transparent AI applications 
in the era of large language models. 
We carefully constructed the \texttt{LPcode} dataset by selecting code under 
Apache, BSD, or MIT licenses, ensuring that all sources meet open and permissive standards for research. 
We remove personally identifiable information, such as email addresses, URLs, and phone numbers, 
from code to safeguard privacy. 
This approach respects the rights of contributors and mitigates ethical risks associated with data collection.
Although this research promotes responsibility and accountability in AI-driven coding, 
a risk of misuse remains. 
Tools such as \texttt{LPcodedec} require careful and balanced application, 
ensuring developers can benefit from AI assistance while safeguarding authorship.


\bibliography{custom}

\appendix
\onecolumn

\section{Dataset Construction: \texttt{LPcode}}
\label{sec:appendix_data_construct}

\subsection{Overview of the Construction Process}

\begin{figure*}[hbt!]
    \centering
        \includegraphics[width=\textwidth]{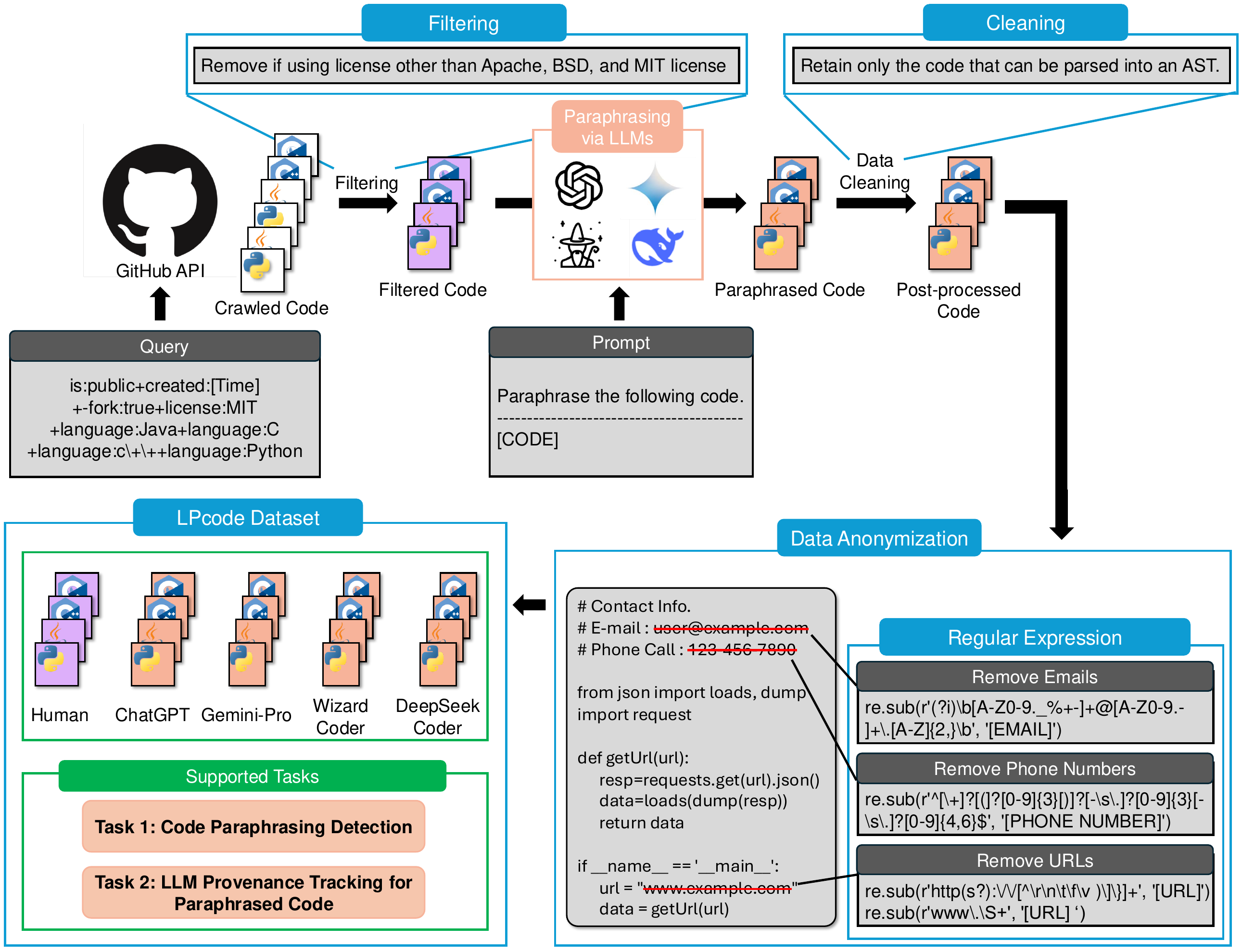}
    \caption{Overview of the \texttt{LPcode} dataset construction process.
    }\label{fig:overview}
\end{figure*}

Figure~\ref{fig:overview} illustrates the process of constructing the \texttt{LPcode} dataset.

\clearpage

\subsection{Number of Code Samples at Each Stage}

\begin{figure*}[hbt!]
    \centering
    \includegraphics[width=\linewidth]{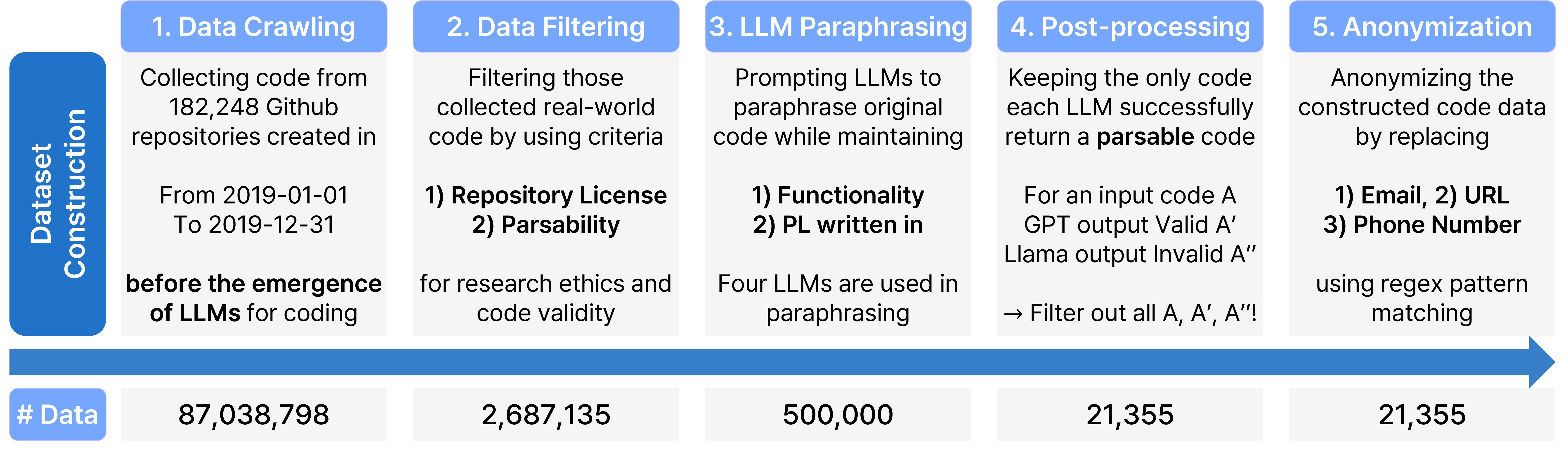}
    \caption{\texttt{LPcode} dataset construction process and the number of code samples at each stage.}\label{fig:detail}
\end{figure*}

Figure~\ref{fig:detail} 
presents the construction process of the \texttt{LPcode} dataset 
along with the number of code samples at each stage.

\paragraph{Data Crawling:} 
Collected C, C++, Java, and Python code from 182,248 GitHub repositories 
created between January 1, 2019, and December 31, 
2019~(a total of 87,038,798 code).

\paragraph{Data Filtering:} 
Selected code with MIT licenses and code parsed into an AST 
to ensure research ethics and code validity~(a total of 2,687,135 code).

\paragraph{LLM Paraphrasing:} Sampled 100,000 code~(25,000 per language) 
and paraphrased them using four LLMs while maintaining functionality 
and the original 
programming language~(400,000 paraphrased code and 100,000 human-written code, 
a total of 500,000 code).

\paragraph{Post-processing:} Retained only human-written code that was 
successfully paraphrased into code that can be parsed into an AST by all four 
LLMs, resulting in a balanced 1:1:1:1:1 ratio~(a total of 21,355 code).

\paragraph{Data Anonymization:} Anonymized the final dataset by removing emails, URLs, 
and phone numbers using regex pattern matching~(a total of 21,355 anonymized code).

\clearpage

\section{Prompt for Code Paraphrasing via LLMs}
\label{sec:appendix_LLM_code_collection}

\begin{figure*}[hbt!]
    \centering
    \includegraphics[width=1\textwidth]{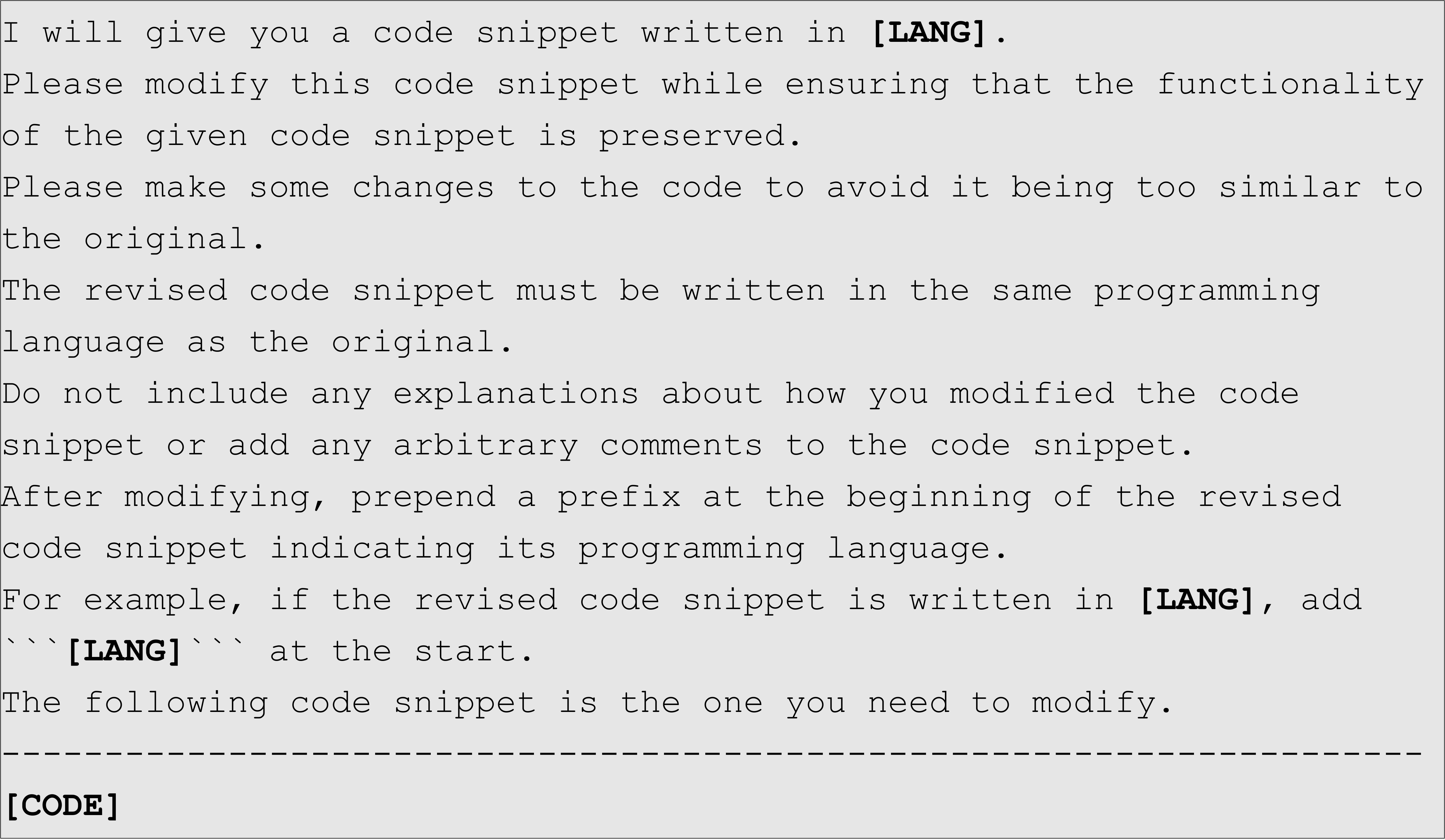}\\
    \caption{Prompt for code paraphrase using LLMs. 
    We utilize LLMs to generate code by incorporating 
    human-written code~(\textbf{[CODE]}) along with 
    their respective programming languages~(\textbf{[LANG]}) into 
    this prompt template.
    }\label{fig:appendix_prompt}
\end{figure*}

Figure~\ref{fig:appendix_prompt} represents the 
prompt used for code paraphrasing with LLMs.
We provide human-written code~(\textbf{[CODE]}) along with their programming languages~(\textbf{[LANG]}), 
and instruct LLMs to paraphrase the code 
while maintaining their functionality and programming language.

\section{Data Statistics}
\label{sec:appendix_data_size}

Table~\ref{tab:data_statistics_task} presents the dataset sizes 
for Task 1 and Task 2.

\begin{table}[hbt!]
\centering

\begin{tabular}{l|cc}

\hline
\noalign{\hrule height 0.8pt}
 & Task 1 & Task 2

\\ 
\hline

C & 3,656 & 1,828 
\\ 
C++ & 3,080 & 1,540
\\ 
Java & 11,952 & 5,976
\\ 
Python & 15,480 & 7,740
\\

\hline
\noalign{\hrule height 0.8pt}

\end{tabular}
\caption{The dataset sizes for Task~1 and Task~2. 
}\label{tab:data_statistics_task}
\end{table}

\clearpage

\section{Coding Style Features}
\label{sec:appendix_ten_features}

Table~\ref{tab:ten_features_detail} presents 
the detailed calculation methods for the 10 coding style features.
The 10 coding style features are grouped as follows:

\paragraph{Naming Consistency:} 1) Function Naming Consistency; 2) Variable Naming Consistency; 3) Class Naming Consistency; and 4) Constant Naming Consistency.

\paragraph{Code Structure:} 1) Indentation Consistency; 2) Function Length; and 
3) Nesting Depth.

\paragraph{Readability:} 1) Comment Ratio; 2) Function Name Length; and 3) Variable Name Length.

\begin{table*}[hbt!]
\centering
\begin{tabular}{l|p{11cm}}
\hline
\textbf{Feature} & \textbf{Description} \\ \hline
Function Naming Consistency 
    & Computes the frequency of each naming convention—camelCase (starts with a lowercase letter and uses capitals at word boundaries), snake\_case (uses underscores between words), PascalCase (starts with an uppercase letter with capitals at word boundaries), UPPER\_SNAKE\_CASE (all uppercase with underscores), and Other—and calculates the ratio of functions following the most common pattern. \\ \hline
Variable Naming Consistency 
    & Measures the uniformity of variable naming by determining the frequency of each naming style and reporting the ratio of the most frequently used pattern to the total number of variables. \\ \hline
Class Naming Consistency 
    & Evaluates how consistently class names follow naming conventions by calculating the ratio of the most common pattern among all class names. \\ \hline
Constant Naming Consistency 
    & Assesses the consistency in naming constants by computing the frequency of each naming style and dividing the count of the most common style by the total number of constants. \\ \hline
Indentation Consistency 
    & Analyzes the uniformity of indentation across the code by determining the frequency of each indentation length and calculating the ratio of the most common indentation pattern relative to all observed indentations. \\ \hline
Function Length 
    & Computes the average number of lines per function, providing insight into the typical size of a function. \\ \hline
Nesting Depth 
    & Measures the average depth of nested code blocks~(such as loops and conditionals) to reflect the overall structural complexity of the code. \\ \hline
Comment Ratio 
    & Calculates the proportion of comment lines relative to the total lines of code, indicating how well the code is documented. \\ \hline
Function Name Length 
    & Computes the average number of characters in function names, offering a measure of how descriptive the function naming is. \\ \hline
Variable Name Length 
    & Determines the average length of variable names in characters, reflecting the level of detail in variable naming. \\ \hline
\end{tabular}
\caption{Detailed descriptions for 10 coding style features.}
\label{tab:ten_features_detail}
\end{table*}

\clearpage

\section{Analysis of Token Frequency Pattern Differences Across LLMs} 
\label{sec:appendix_tfidf}

\begin{figure*}[hbt!]
    \centering
    \subcaptionbox{C}{
    \includegraphics[width=0.24\textwidth]{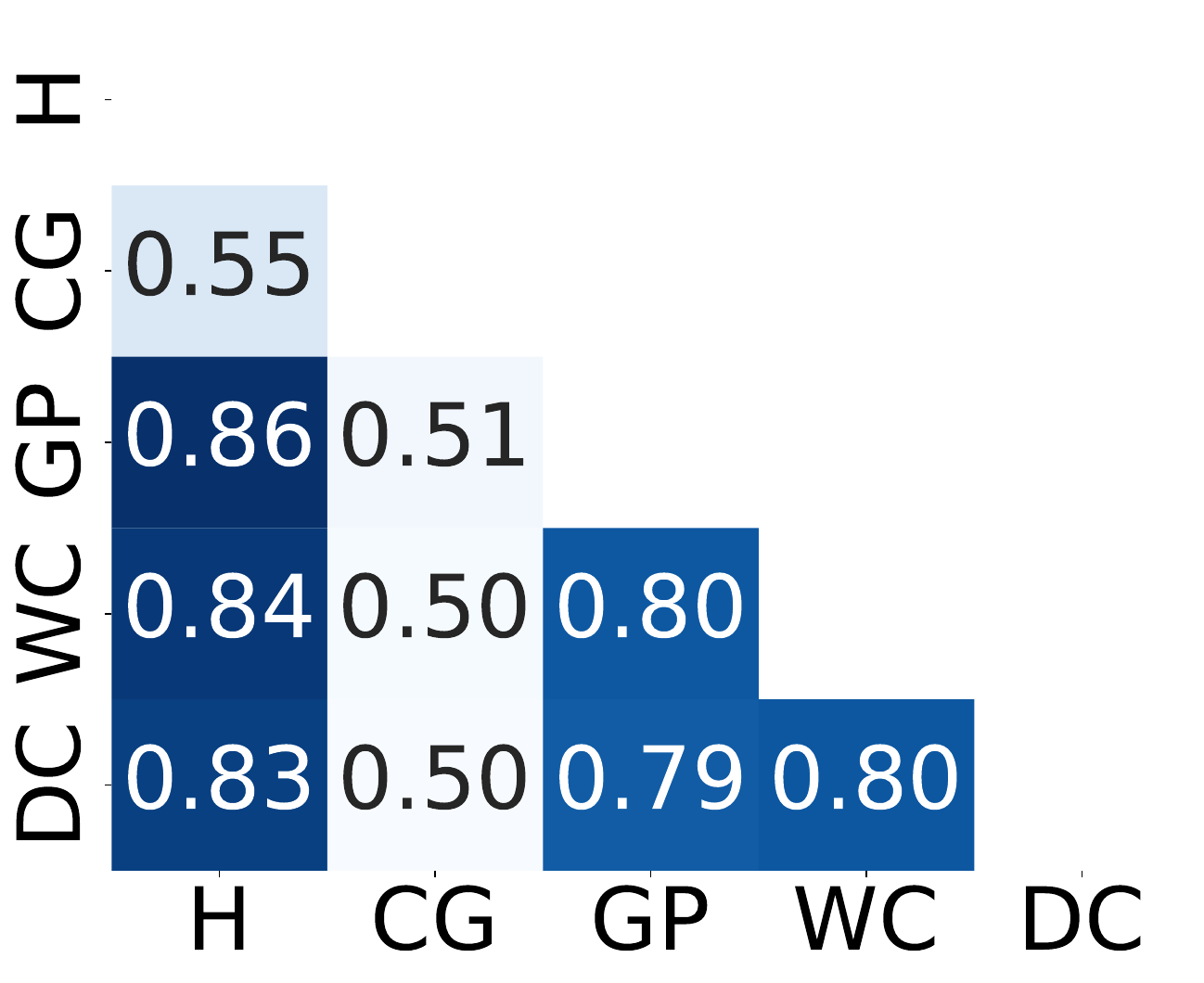}}
    \subcaptionbox{C++}{
    \includegraphics[width=0.24\textwidth]{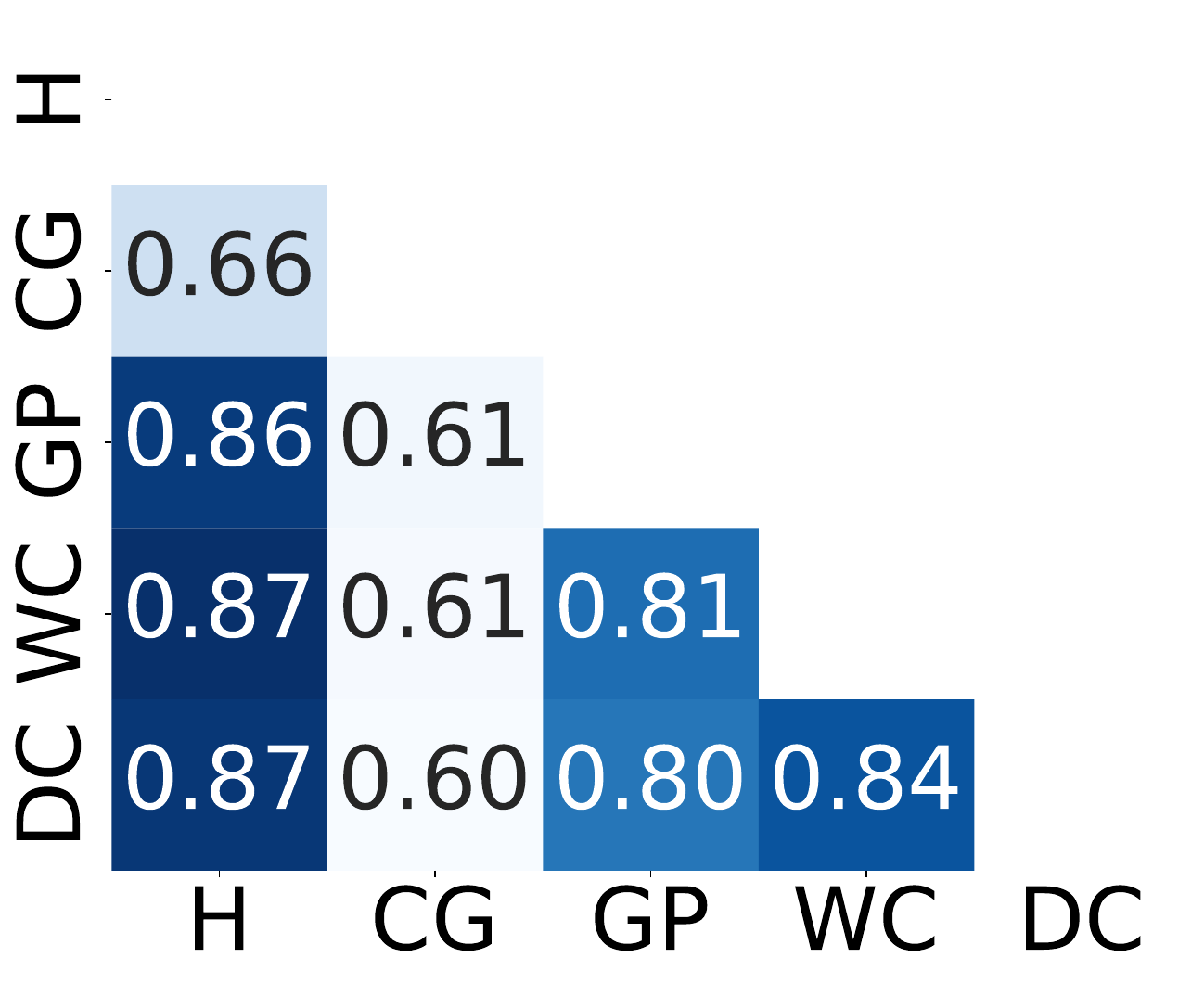}}
    \subcaptionbox{Java}{
    \includegraphics[width=0.24\textwidth]{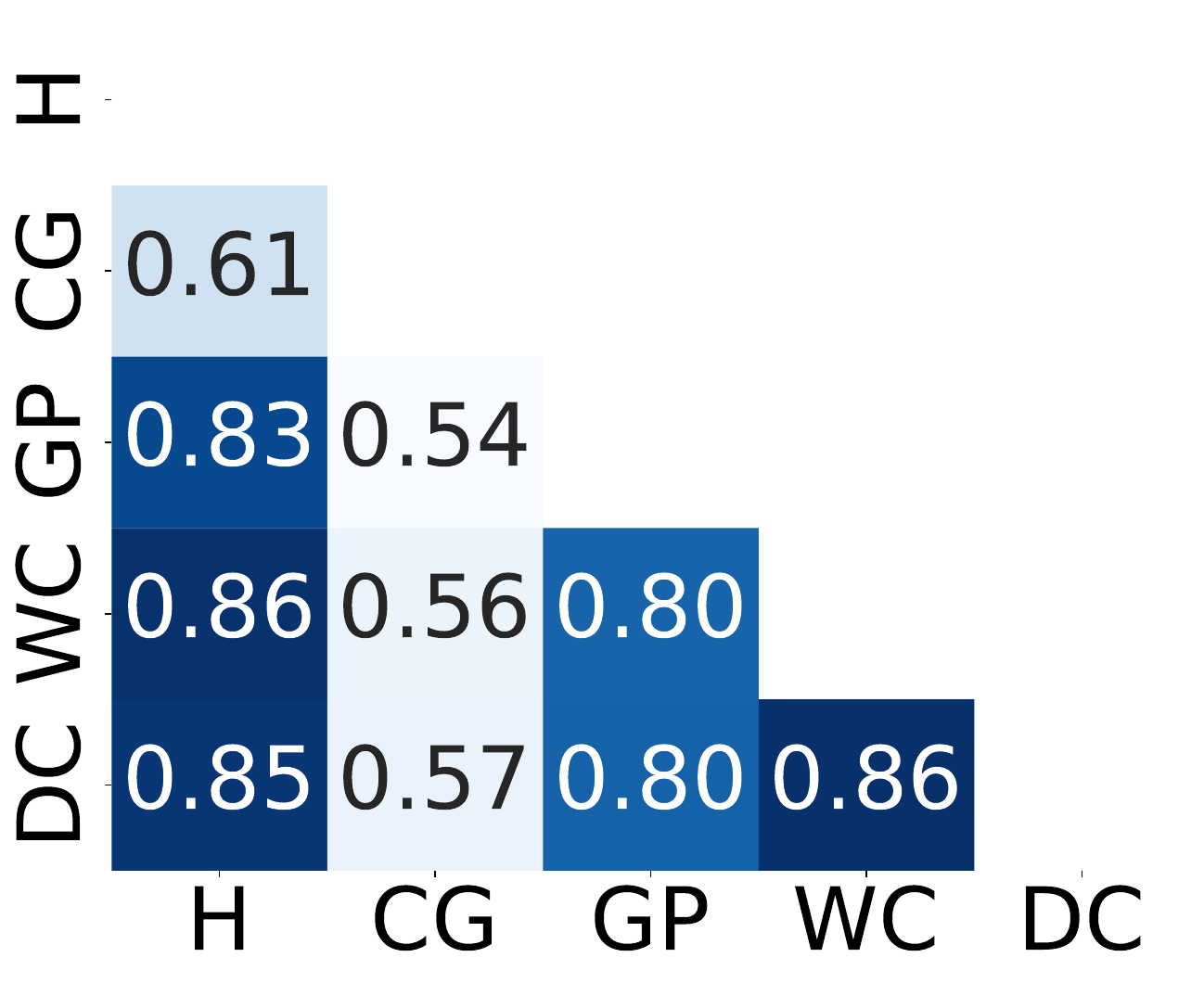}}
    \subcaptionbox{Python}{
    \includegraphics[width=0.24\textwidth]{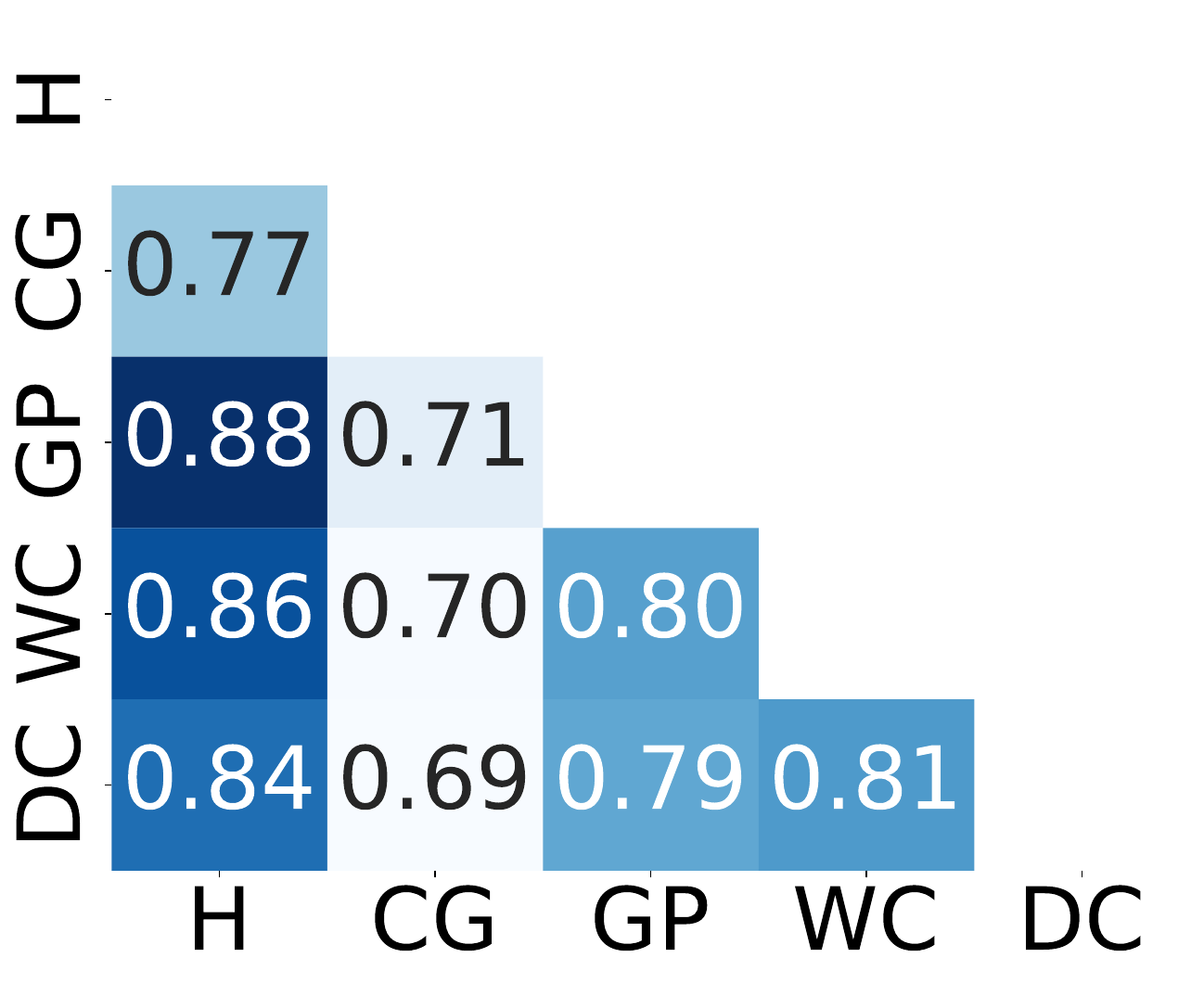}}
    \caption{A heatmap illustrating the pairwise 
    cosine similarity of TF-IDF vectors 
    computed from code samples generated by 
    different sources: 
    Human~(H), ChatGPT~(CG), Gemini-Pro~(GP), 
    WizardCoder~(WC), and DeepSeek-Coder~(DC).
    Darker cells in the heatmap indicate more similar 
    token frequency patterns between the two generators.
    The diagonal elements represent the self-similarity 
    of each generator, which is expectedly 1.00.
    }\label{fig:tfidf} 
\end{figure*}

The heatmap in Figure~\ref{fig:tfidf} 
shows the token frequency patterns of 
code samples written by humans and four LLMs, 
calculated using TF-IDF vectors and measured with 
cosine similarity. 
A higher cosine similarity value means the 
two generators use tokens in a more similar way. 
In all four programming languages, 
ChatGPT consistently shows lower similarity with 
the other LLMs, 
suggesting that its token usage patterns are 
different from those of the others. 
At the same time, ChatGPT has relatively higher 
similarity with human-written code compared 
to the other LLMs, which indicates that ChatGPT 
tends to generate code that is more similar 
to human style. 
In contrast, Gemini-Pro, WizardCoder, and 
DeepSeek-Coder share similar token usage patterns 
with each other.

\clearpage

\section{Similarity Analysis Between an Original Code and its Paraphrased Code}
\label{sec:codebleu_ana}

We compare and analyze the similarity between  and 
its paraphrased versions generated by LLMs using the CodeBLEU metric. 
CodeBLEU is an automated evaluation metric that 
incorporates programming language-specific characteristics, 
providing a more suitable assessment compared to BLEU.
CodeBLEU is computed as a weighted combination of four components: 
1) BLEU: Token-level n-gram similarity. 
2) Weighted n-gram match: Assigns higher weights to important programming language keywords. 
3) Syntactic AST match: Compares the structural similarity of code using AST subtree matching.
4) Semantic data-flow match: Analyzes the data flow to assess semantic similarity, 
capturing variable dependencies. 

\begin{figure*}[hbt!]
    \centering
    \subcaptionbox{C}{
    \includegraphics[width=0.24\textwidth]{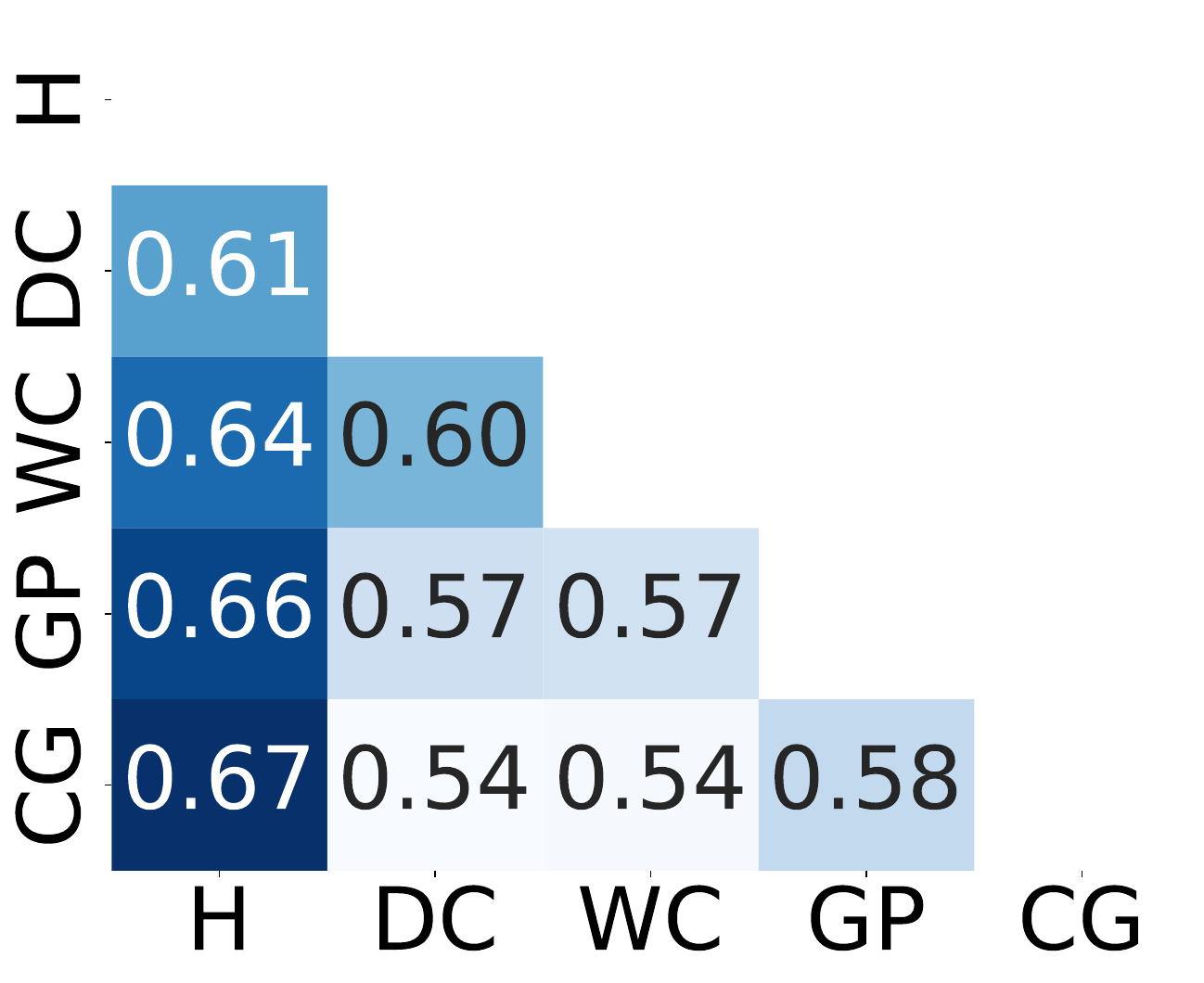}}
    \subcaptionbox{C++}{
    \includegraphics[width=0.24\textwidth]{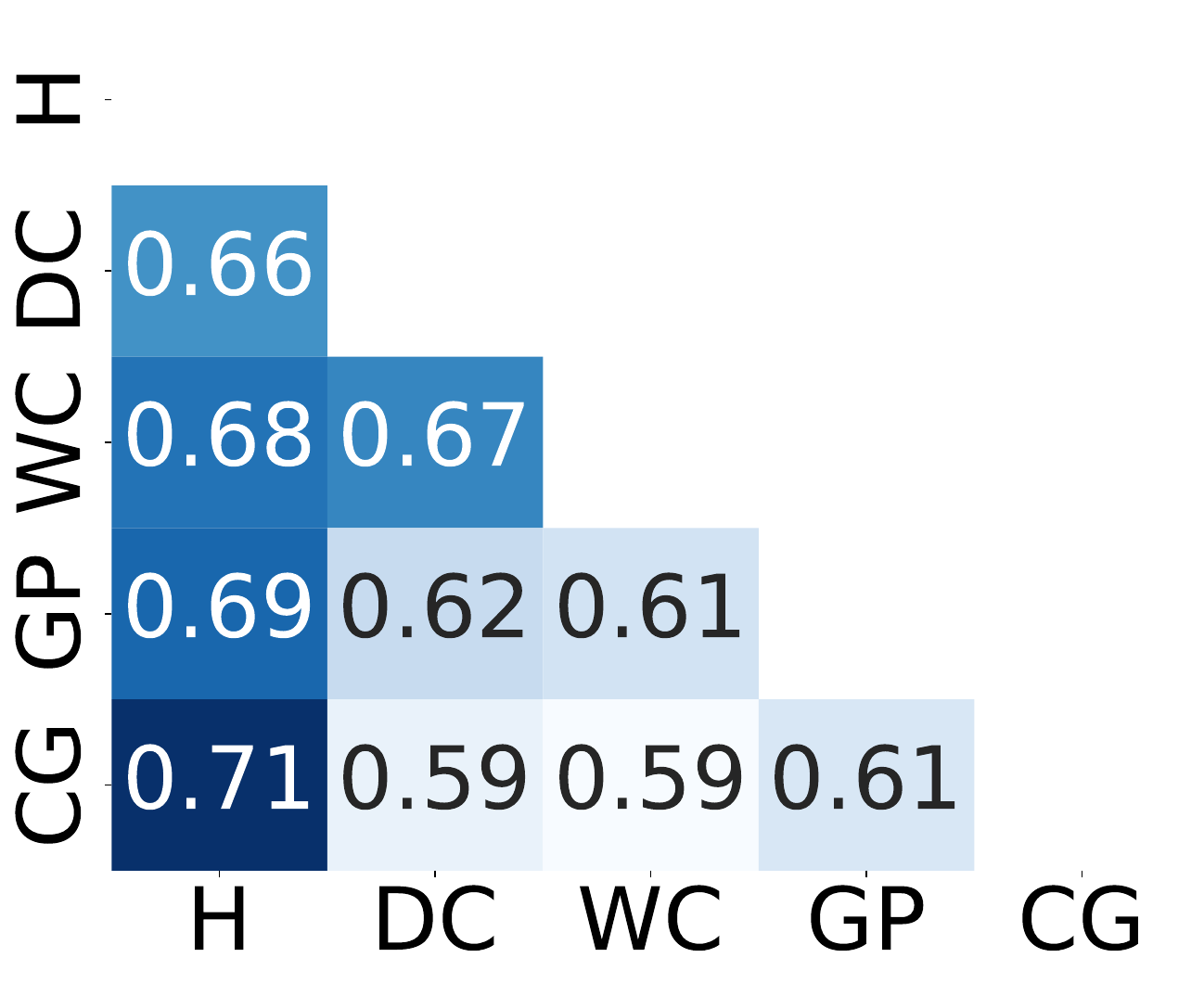}}
    \subcaptionbox{Java}{
    \includegraphics[width=0.24\textwidth]{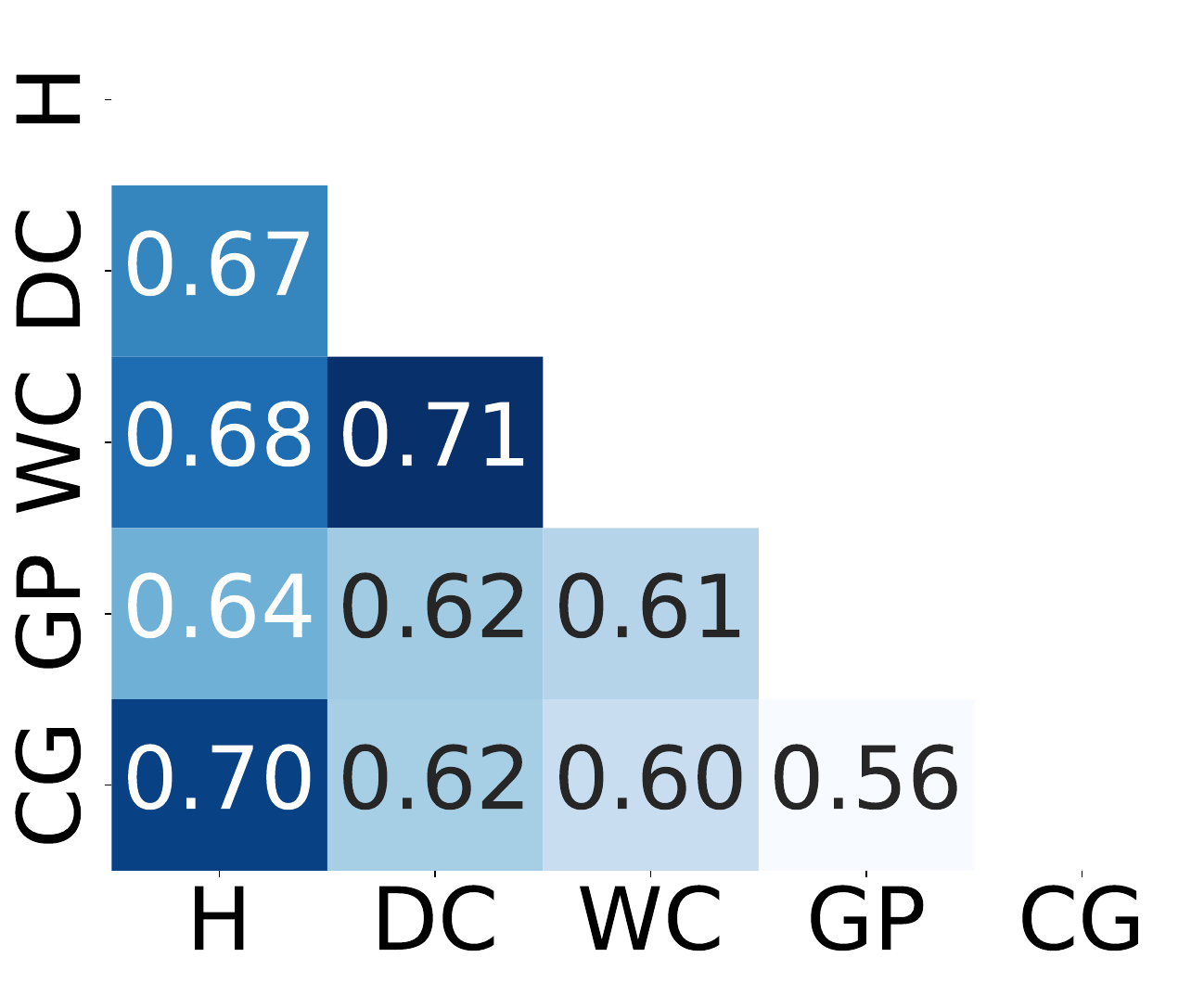}}
    \subcaptionbox{Python}{
    \includegraphics[width=0.24\textwidth]{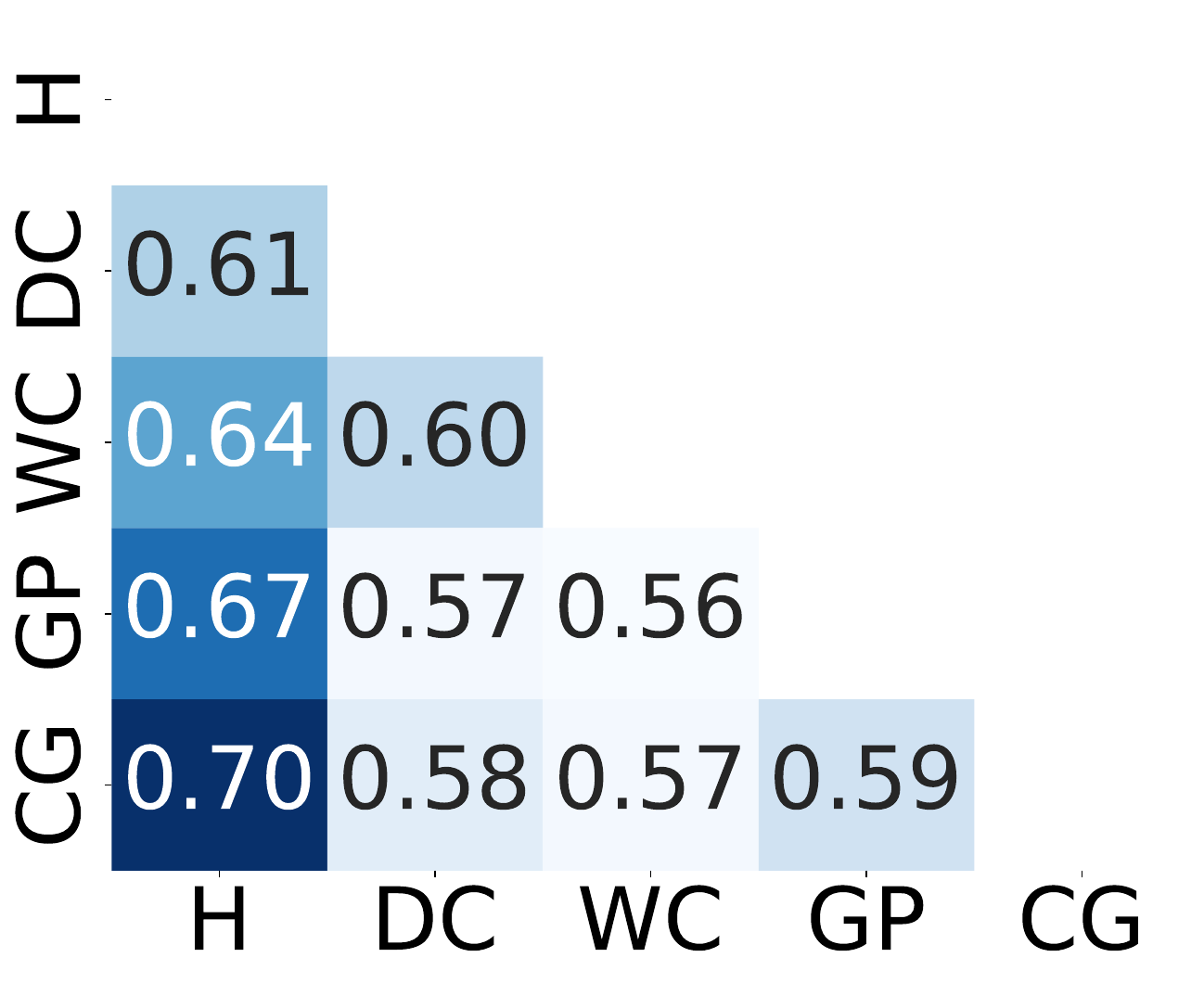}}
    \caption{A heatmap illustrating the average CodeBLEU scores, 
    highlighting both the similarity 
    between human-written code and its paraphrased versions generated by four LLMs, 
    as well as the relationships among the paraphrased code produced by different LLMs.
    H denotes human-written code, 
    while DC, WC, GP, and CG represent code paraphrased 
    by DeepSeek-Coder, WizardCoder, Gemini-Pro, and ChatGPT, respectively.
    }\label{fig:codebleu} 
\end{figure*}

Figure~\ref{fig:codebleu}
is a heatmap visualizing the average CodeBLEU scores, 
showing both the similarity between human-written code and the paraphrased versions 
generated by four LLMs, 
as well as the similarity among the paraphrased code produced by different LLMs.
We observe that, overall, the code paraphrased by 
commercial LLMs~(ChatGPT and Gemini-Pro) tends to have a higher similarity to the 
human-written original code compared to the code paraphrased by 
open-source LLMs~(WizardCoder and DeepSeek-Coder).
Commercial LLMs are likely to perform paraphrasing in a way that preserves 
a more human-friendly coding style, 
as they have a larger parameter size compared to open-source LLMs 
and are trained on more extensive datasets using reinforcement learning techniques such as 
RLHF~(Reinforcement Learning from Human Feedback).

With CodeBLEU scores mostly below 0.7 between human-written code and its paraphrased versions 
by LLMs, these paraphrases can be categorized as 
moderately Type-3 clones~(syntactically modified)
based on the code clone classification defined by \citet{svajlenko2014towards}.
\citet{svajlenko2014towards} 
established a clear boundary between 
Type-3 and Type-4 clones and instead categorized them into three groups based on similarity values.
Specifically, Strongly Type-3 clones fall within the [0.7, 1.0) similarity range, 
maintaining a high resemblance to the original code while incorporating minor modifications 
such as variable renaming or simple syntax changes. 
Moderately Type-3 clones have a similarity range of [0.5, 0.7), 
exhibiting increased structural differences due to modifications in control flow 
or multiple code block alterations. 
Lastly, Weakly Type-3 + Type-4 clones fall within the [0.0, 0.5) similarity range, 
featuring significantly different structures from the original code, 
making the boundary between Type-3 and Type-4 clones ambiguous.

The lower CodeBLEU scores between code paraphrased by different LLMs, 
compared to those between human-written code and its LLM-paraphrased versions, 
suggest that LLMs vary in their approaches to paraphrasing human-written code.
This supports the necessity of the task of tracking the specific LLM 
that paraphrased human-written code, 
which is one of the problems we aim to address.
Among the LLMs, DeepSeek-Coder and WizardCoder show the highest CodeBLEU scores 
between their paraphrased code, 
indicating that open-source LLMs tend to follow a similar approach 
when paraphrasing human-written code.
These results suggest that identifying which LLM paraphrased 
human-written code may be more challenging among open-source LLMs 
than commercial ones.

\end{document}